\documentclass[fleqn,10pt]{wlscirep}
\usepackage[utf8]{inputenc}
\usepackage[T1]{fontenc}
\title{From driving automation systems to autonomous vehicles: clarifying the terminology}

\author[1,2]{David Fern\'andez Llorca}
\affil[1]{Computer Engineering Department, University of Alcal\'a, Alcal\'a de Henares, Spain.}
\affil[2]{European Commission, Joint Research Center, Seville, Spain 
\newline Email: david.fernandez-llorca@ec.europa.eu}

\begin{abstract}
The terminological landscape is rather cluttered when referring to autonomous driving or vehicles. A plethora of terms are used interchangeably, leading to misuse and confusion. With its technological, social and legal progress, it is increasingly imperative to establish a clear terminology that allows each concept to be placed in its corresponding place. 
\end{abstract}

\begin{document}

\flushbottom
\maketitle
%
%
\thispagestyle{empty}

\section*{}
\emph{Autonomous}, \emph{automated}, \emph{self-driving}, \emph{robotic}, \emph{driverless}, \emph{connected}, \emph{cooperative} - all these terms are commonly used interchangeably, and not only in the media and society \cite{Bigelow2019}, but also in industry and academia. As progress is made in this field, at the scientific-technological, social (user acceptance) and legal levels, and accelerated by current needs \cite{Llorca2020}, there is less time left to establish an effective terminology that allows each concept to be unequivocally identified. The main issue, which generates most of the confusion, and legal uncertainty, arises when it comes to clarifying the role of humans, who can be considered as \emph{assisted} drivers, \emph{assistant} or \emph{backup} drivers, or mere \emph{passengers} without responsibility on the driving tasks. Conceptual problems and misunderstandings in the way we name autonomous driving systems make it difficult for users, both those using the vehicle and those interacting externally with it, to understand the technology and thus to adopt it.  Despite all its limitations, clarification of terminology involves addressing the automated driving tasks and the levels of automation, and relate them with the role of the users. 

\section*{Automated driving tasks}
\begin{figure}[ht]
\centering
\includegraphics[width=0.8\linewidth]{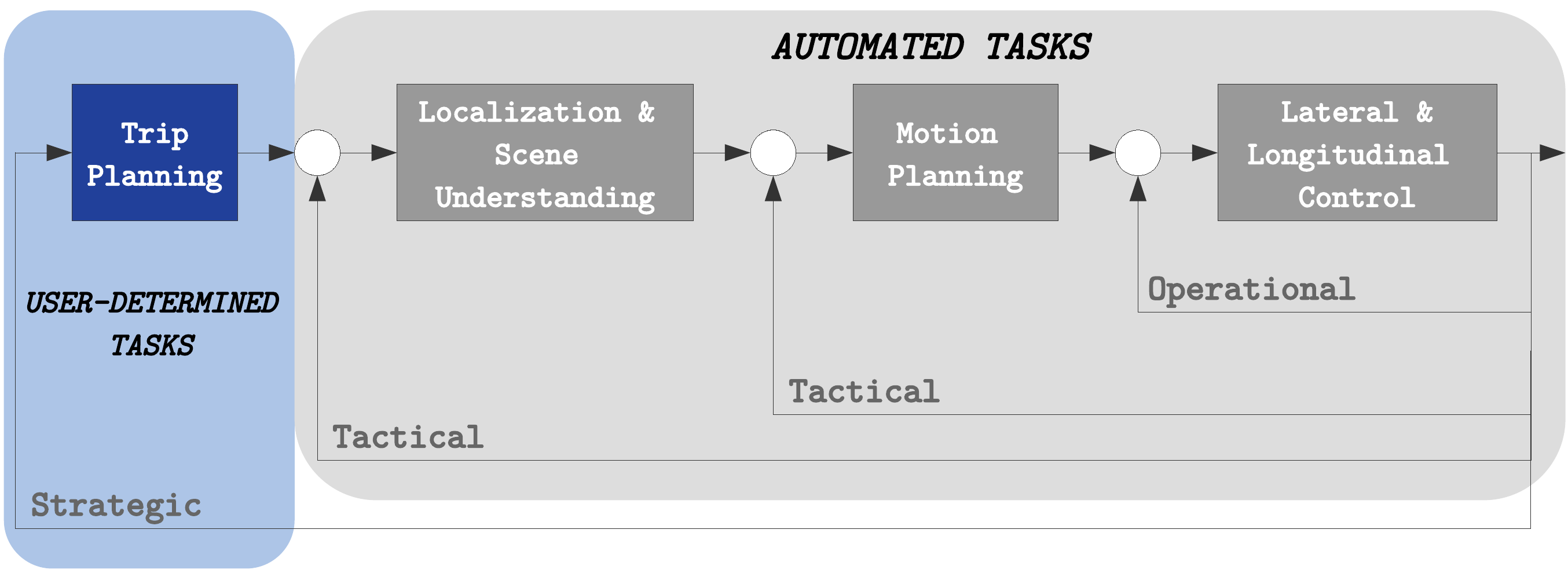}
\caption{Schematic view of user-determined (strategic) and automated (tactical and operational) driving tasks.}
\label{fig:1}
\end{figure}

The overall act of driving can be mainly divided into three driving tasks \cite{Michon1985} (see Fig. \ref{fig:1}). First, \emph{strategic} tasks, which involve the global planning of the trip. The control of these tasks is user-determined, either by direct interaction or by defining the objectives of a possible automation. Second, \emph{tactical} tasks, which involves the high-level manoeuvring of the vehicle in traffic during a trip constrained by the directly prevailing local circumstances. When automated, these tasks require dealing with perception (vehicle localization and scene understanding) and motion planning problems. Finally, \emph{operational} tasks, which represent the low-level control actions, such as to stabilize to the reference path or to avoid obstacles or hazardous events. Automation at this level involves solving lateral and longitudinal control problems. 

\section*{Levels of automation}
Although the recommended practice J3016 \cite{SAE_2018} of SAE International is not devised as a technical specification and imposes no requirements, it can be considered a "de facto" standard, as it is increasingly being adopted globally and applied in more and more legal frameworks. It establishes six levels of driving automation, from 0 (no automation) to 5 (full automation). 

The core of the automation is based on the \emph{Driving Automation Systems} (also known as \emph{Automatic Driving Systems}, ADS, when referring to high levels of automation), which are defined as the hardware and software that are capable of performing part or all of the operational and tactical driving tasks (strategic tasks are excluded). A common misconception is that the levels apply to the whole vehicle. However, they only apply to certain driving automation systems, which are subsystems of the vehicle.  Another important concept is the \emph{Operational Design Domain (ODD)} which refers to operating conditions under which a given driving automation system is specifically designed to function. For the purpose of self-containment, we briefly summarize the six levels of automation:

\begin{itemize}
    \item \emph{Level 0 - No Driving Automation}: the performance by the driver of all driving tasks.
    \item \emph{Level 1 - Driver Assistance}: the sustained and (limited) ODD-specific execution by a Driving Automation System of the lateral or the longitudinal vehicle motion control tasks (but not both simultaneously) with the expectation that the driver will perform the remaining driving tasks.  
    \item \emph{Level 2 - Partial Driving Automation}: the sustained and (limited) ODD-specific execution by a Driving Automation System of \emph{both} the lateral and longitudinal vehicle motion control tasks, with the expectation that the driver completes the remaining driving tasks and supervises the Driving Automation System.
    \item \emph{Level 3 - Conditional Driving Automation}: the sustained and (limited) ODD-specific performance by a Driving Automation System of all tactical and operational driving tasks with the expectation that the driver is receptive to a requests to intervene in case of system failures, and will respond appropriately. 
    \item \emph{Level 4 - High Driving Automation}: the sustained and (limited) ODD-specific performance by a Driving Automation System of all tactical and operational driving tasks without any expectation that a user will respond to a request to intervene. The Driving Automation System must be capable of reaching a minimal risk condition in the event where the ODD limit is being reached.
    \item \emph{Level 5 - Full Driving Automation}: the sustained and unconditional (i.e., not ODD-specific) performance by a Driving Automation System of all tactical and operational driving tasks without any expectation that a user will respond to a request to intervene.   
\end{itemize}

\begin{figure}[ht]
\centering
\includegraphics[width=0.8\linewidth]{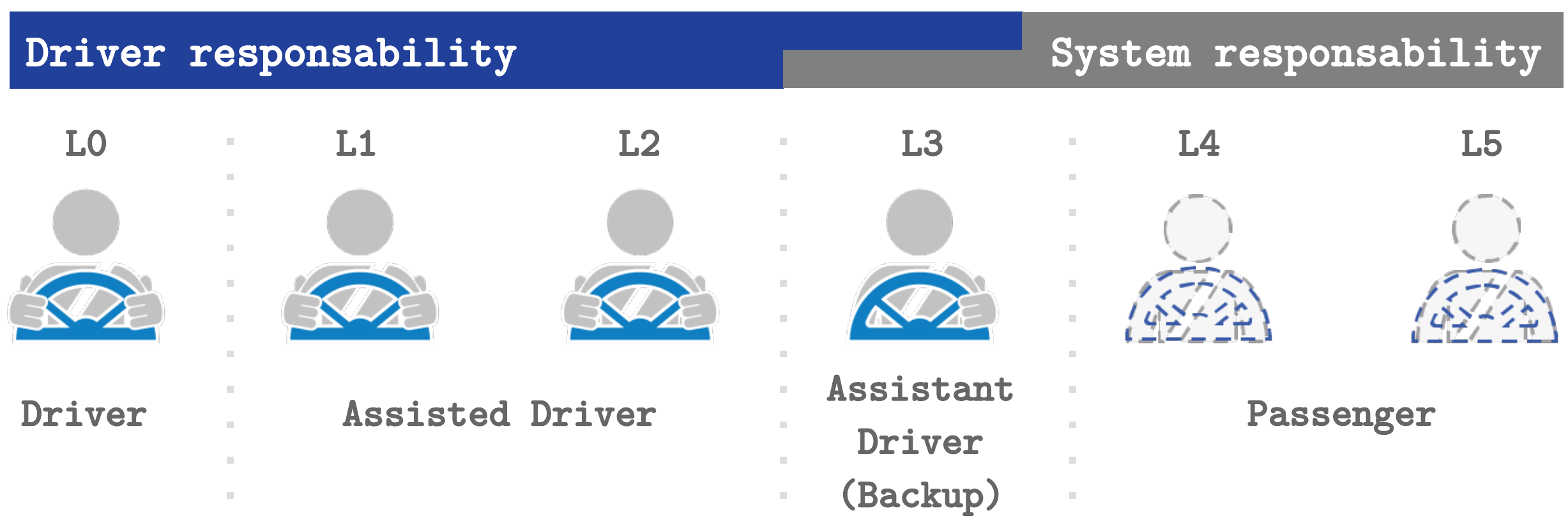}
\caption{Driver's role and responsibility for SAE automation levels.}
\label{fig:2}
\end{figure}

The SAE taxonomy has some well-known problems \cite{Inagaki2019} such as the undefined requirements for safe transition from one level of automation to another, the question of what should happen when the driver does not respond to a request to intervene in level 3, or what is a minimum acceptable risk condition, among others. Despite reasonable criticism \cite{Templeton2018}, and those who believe we should start from scratch \cite{Roy2018}, the SAE levels of automation is a very solid basis and it makes more sense to further improve and adapt it. An interesting approach would be to clarify the role and responsibility of humans in the use of automation systems (see Fig. \ref{fig:2}). Thus, for example, levels 1 and 2 define systems that assist the driver, at level 3 it is the driver who ultimately assists the system, and finally, at levels 4 and 5, the driver is a mere passenger responsible only for strategic tasks. Humans are fully responsible for levels 1 and 2, shared responsibility should be considered in level 3, while no responsibility can be attributed to them for levels 4 and 5. In this respect, level 3 is rather confusing, and can be seen as a perhaps necessary, but not desirable, transitional step towards higher levels of automation.

\section*{Terminology}
\begin{figure}[ht]
\centering
\includegraphics[width=0.6\linewidth]{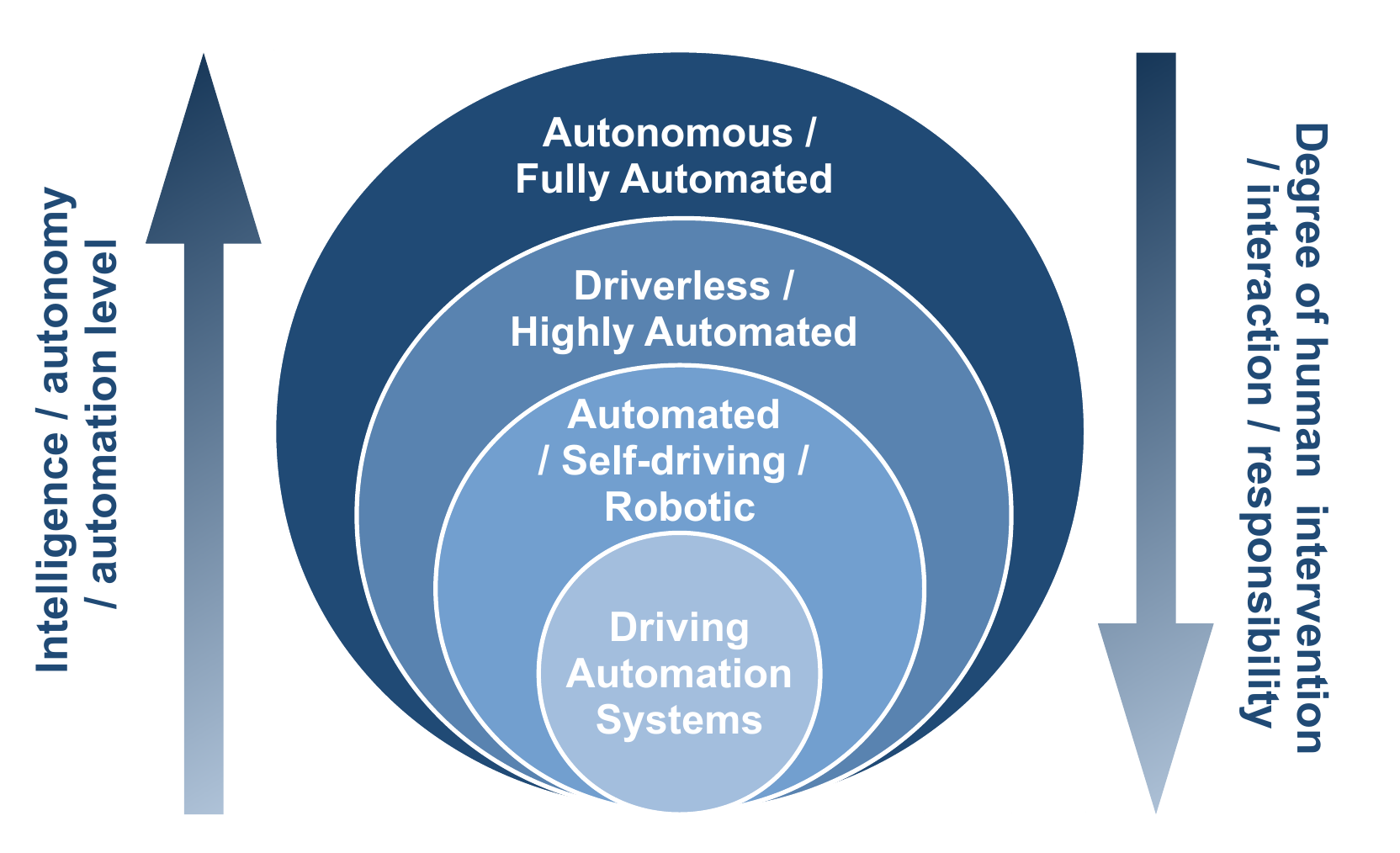}
\caption{Proposed Venn diagram to illustrate how the different terms used when referring to autonomous vehicles are related.}
\label{fig:venn}
\end{figure}

In order to clarify and relate the plethora of terms we propose the Venn diagram depicted in Fig. \ref{fig:venn}. From a bottom-up perspective we have an increased level of automation, intelligence and autonomy, as well as a decreased level of human intervention, interaction and responsibility. Following the approach established by SAE International standard, we place Driving Automation Systems as the core concept from which we can derive the rest. The proposed diagram makes certain assumptions. First, the terminology can focus either on the \emph{vehicle} as a whole or on the \emph{driving} automation features. Second, it is assumed that higher levels of automation implicitly contain lower levels of automation, for instance, a fully automated vehicle (level 5) will also have the ability to operate at lower levels of automation (levels 1 to 4). Third, high levels of automation are considered to require a certain level of connectivity to ensure their operation. Finally, the fact that two terms appear at the same level does not imply that they are completely equivalent. In general we can state that all terms are to some extent imprecise and require a certain level of additional context.

\subsection*{Automated}
This is the most commonly accepted term to characterize vehicles equipped with driving automated systems (i.e., \emph{automated} vehicles or driving systems). However, the main disadvantage is that it does not clearly define the level of automation. When we say \emph{automated} we do not know if we are referring to a low or a high automation level. Therefore, some context, such as "partially", "conditionally", "highly" or "totally", must be added to clearly identify the concept. If this context is not provided, we can fairly consider that \emph{automated} refers to a medium level of automation, as a compromise solution. 

\emph{Automated} is equivalent to \emph{automatic}. The definition of the term \emph{automatic} includes statements such as "that does not require an operator" and "that works by itself under fixed conditions, with little or no direct human control". The fact that automated vehicles are able to perform some driving automated function by themselves is one of the reasons to consider \emph{automated} equivalent to \emph{self-driving}. As an action or thing, \emph{automatic} is also defined as "self-generated" and "self-acting". Thus, the argument used to indicate that the term \emph{autonomous} is inadequate, concerning its capacity for "self-government", is also applicable to some extent to the term \emph{automated}. However, as will be described when addressing the term \emph{autonomous}, this criticism does not clearly hold. 

\subsection*{Robotic}
This term is usually applied to refer to autonomous vehicles as \emph{robocars} or \emph{robotaxis}. It is equivalent to \emph{automated} and although it is used colloquially to connote high levels of automation, it shares the same problem, i.e., it does not allow to establish the specific level of automation.

\subsection*{Self-driving}
This term has become synonymous with \emph{automated}. \emph{Self-driving} clearly identifies that the vehicle or the system is performing the driving tasks by itself. But, as it happens with \emph{automated}, the main disadvantage is that \emph{self-driving} does not clarify the level of automation of the driving feature or the vehicle. Therefore, \emph{self-driving} does not necessary mean that no driver or user is present. 

\subsection*{Connected}
The term \emph{connected} is usually considered as an additional feature of the driving automation systems. There is a widespread trend in scientific, industrial and political circles to speak of \emph{"Connected and Autonomous Vehicles"} or \emph{"Connected and Automated Vehicles"} (the widely used acronym, CAV, or AV, applies to both \emph{autonomous} and \emph{automated} vehicles). This is reasonable to some extent since \emph{connected} does not necessary mean \emph{automated} (we can have fully manual connected cars), hence the tendency to use the term \emph{connected} explicitly. Actually, in-vehicle connectivity is continuously increasing (e.g., by using integrated modems with a SIM card), in part because of mandatory e-call \cite{Ciuffo2017}, but also because of new potential services (e.g., software updates, updated traffic state for navigators, data recording, etc.) and Internet access for driver and passengers.  

Therefore, automated vehicles can complementary be connected to other vehicles (V2V), to the infrastructure (V2I), to vulnerable road users (V2VRUs) and to the network (V2N). In the case of having connectivity to all of the above, we call it vehicle-to-everything (V2X) connectivity. Connectivity relies on different technologies, including dedicated short range communication (i.e., IEEE 802.11p) for V2V, V2I and V2VRUs, and cellular V2X (from 3GPP to 5G NR C-V2X). The advantages of high-reliability, low-latency connectivity to enhance the intelligence and automation level of vehicles are obvious. V2X provides 360 degree, non-line of sight sensing with higher ranges than onboard sensors such as cameras, radar or LiDAR \cite{Parra2019}, allows sensor sharing between vehicles and real-time updates from the infrastructure, increases situation awareness, and makes it possible to perform predictive and coordinated driving by exchanging intentions and sensor data \cite{Qualcomm2019}. These benefits clearly enhance perception capabilities, intelligence, autonomy and automation level of the driving automation systems, so higher levels of automation are expected to require connectivity. 

\subsection*{Cooperative}
It is common to associate \emph{connected} with \emph{cooperative}. This is because cooperative systems (whether manual or autonomous/automated) require connectivity between agents and infrastructure (although it is possible to cooperate with non-connected vehicles \cite{Parra2018}, connectivity is a fundamental enabler for cooperation). However, it is important to note that \emph{connected} does not necessary mean \emph{cooperative}. We can have \emph{connected} and \emph{automated} vehicles that take advantage of the aforementioned benefits of connectivity without performing cooperative driving. 

Cooperation is considered here as an additional feature of the driving automation system that affects the individual operational, tactical, and even strategic, tasks, with the objective of obtaining individual behaviours subject to the optimisation of the collaborative behaviour of all agents involved. This feature may or may not be present at each level of automation. 

\subsection*{Driverless (or unmanned)}
This term makes explicit the fact that the automated vehicle has no (backup) driver. Therefore, the vehicle can be driving empty ("personless") or consider all users in the vehicle as passengers. Therefore, a \emph{driverless} (or \emph{unmanned}) vehicle refers to a high level of automation in which there is no expectation that a user will respond to a request for intervention. However, the term \emph{driverless} by itself does not sufficiently differentiate between highly automated (level 4) or fully automated (level 5). Additional context information is required to know whether the ODD is limited or unconditional. This is why the location of \emph{driverless} in the proposed Venn diagram is at the same level as highly automated.

In the SAE recommended practice it is remarked that this term does not clarify if a vehicle is remotely operated by a human driver. However, the fact that the vehicle is operated by a remote driver should be given the same consideration as if it were driven by a conventional driver, without altering the level of automation (from 0 to 3). Therefore, when referring to \emph{autonomous} or \emph{automated} vehicles as \emph{driverless} vehicles, there should be no confusion about the possible presence of a remote driver. 

\subsection*{Autonomous}
As stated by the SAE recommendation, the term \emph{autonomous} "\emph{has been used for a long time in the robotics and artificial intelligence research communities to signify systems that have the ability and authority to make decisions independently and self-sufficiently}" (the definition of \emph{autonomy} refers to \emph{self-governance} or \emph{independence}). Over time, it has become a synonymous with \emph{automated} since its use "\emph{was casually broadened to not only encompass decision making, but to represent the entire system functionality}". 

It is true that the exact definition of the word \emph{autonomy} is not of great help in its application to the field of automation. For example, it seems more complicated to establish "levels of autonomy" than "levels of automation". However, its use is becoming more and more widespread. As described in \cite{Zhang2020}, "\emph{most AV developers prefer the term autonomous vehicle, to emphasize the vehicle-borne intelligence}". For a system (whether software only or embedded in a robot or vehicle) to be \emph{autonomous} requires a high level of intelligence and sophistication. When we talk about \emph{autonomous} vehicles or driving features, we are talking about the highest levels of automation. The idea that the system should be self-sufficiently makes us think of an unconditional ODD. If we also consider that the higher levels of automation are capable of implementing lower levels, the term \emph{autonomous} is placed at the highest level of the hierarchy, equivalent to \emph{fully automated} (level 5). Therefore, whereas \emph{automated} does not clarify the level of automation, \emph{autonomous} directly refers to highest levels. 

As for the criticism that establishes \emph{connected} and \emph{cooperative} as opposite to \emph{autonomous} the following can be stated. First, connectivity can be considered as an additional input that enhances perception. Receiving information from other agents, or from the infrastructure, with V2X communications does not imply losing the autonomous nature. For example, from the point of view of the level of automation or the autonomy, it makes no difference whether the status of a traffic light is obtained by a wireless connection or by a vision-based recognition system, or whether the intention of vehicles to change lanes is sent with V2V communications or estimated using on-board sensors and advanced artificial intelligence systems. Second, \emph{cooperative} refers to coordinated and collaborative objectives which can affect to all individual driving tasks (operational, tactical and strategic) of each vehicle. It is assumed that these shared objectives, that guide the behaviour of each individual vehicle, would lead to a higher level of automation and intelligence, and therefore would not negatively affect the degree of autonomy. We can even consider this case as an autonomous fleet. If cooperation brings benefits, an autonomous vehicle will always tend to behave cooperatively rather than isolated. In other words, \emph{autonomous} does not mean \emph{isolated}.

Another common criticism with the use of this term is related with its \emph{self-governance} nature. A well known comment (circa 2012, whose origin is somewhat uncertain), to support why a vehicle could never be autonomous as this implies self-governing said: \emph{"If my vehicle truly was autonomous, then I would say "Take me to work" and it would take me to the beach instead"} \cite{Godsmark2017}. Actually, the SAE recommendation states that "\emph{even the most advanced ADSs are not self-governing. Rather, ADSs operate based on algorithms and otherwise obey the commands of users}". However, the answer to this critic statement can be found in the SAE document itself, when referring to the possible automation of strategic aspects of driving: "\emph{Strategic aspects of vehicle operation (decisions regarding whether, when, and where to go, as well as how to get there) are excluded from the definition of driving tasks, because they are considered user-determined aspects of the broader driving task. However, for certain advanced ADS applications, such as some ADS-dedicated vehicle applications, timing, route planning and even destination selection may also be automated in accordance with purposes defined by the user}". That is, when we talk about vehicle automation levels, we are referring to the operational and tactical aspects of driving, which also applies to the concept of \emph{autonomous}. The strategic tasks will always be user-determined. Therefore, a \emph{fully automated} or \emph{autonomous} vehicle will never decide against user commands. In other words, \emph{it will never decide to take us to the beach if we have asked it to take us to work} . 

\vspace{8mm}

There is still a long way to go in the conceptual revisions, including the introduction of new terms such as \emph{autonowashing} \cite{Dixon2020} or \emph{geotonomous/geotonomy} \cite{Roy2018b}, as we move towards the adoption of autonomous vehicles. In the meantime, we humbly hope that this attempt to clarify terminology will be as useful as possible. 

\bibliography{sample}

\end{document}